# Face Recognition Based on Polar Frequency Features

YOSSI ZANA[1]

Dept. of Computer Science – IME, University of São Paulo, São Paulo - SP, Brazil

and

ROBERTO M. CESAR-JR

Dept. of Computer Science – IME, University of São Paulo, São Paulo - SP, Brazil

___

A novel biologically motivated face recognition algorithm based on polar frequency is presented. Polar frequency descriptors are extracted from face images by Fourier-Bessel transform (FBT). Next, the Euclidean distance between all images is computed and each image is now represented by its dissimilarity to the other images. A Pseudo-Fisher Linear Discriminant was built on this dissimilarity space. The performance of Discrete Fourier transform (DFT) descriptors, and a combination of both feature types was also evaluated. The algorithms were tested on a 40- and 1196-subjects face database (ORL and FERET, respectively). With 5 images per subject in the training and test datasets, error rate on the ORL database was 3.8, 1.25 and 0.2% for the FBT, DFT, and the combined classifier, respectively, as compared to 2.6% achieved by the best previous algorithm. The most informative polar frequency features were concentrated at low-to-medium angular frequencies coupled to low radial frequencies. On the FERET database, where an affine normalization pre-processing was applied, the FBT algorithm outperformed only the PCA in a rank recognition test. However, it achieved performance comparable to state-of-the-art methods when evaluated by verification tests. These results indicate the high informative value of the polar frequency content of face images in relation to recognition and verification tasks, and that the Cartesian frequency content can complement information about the subjects' identity, but possibly only when the images are not pre-normalized. Possible implications for human face recognition are discussed.

___

## 1. INTRODUCTION

Face recognition is a highly complex task due to the many possible variations of the same subject in different conditions, like luminance and facial expressions, and the three-dimensional nature of the head. Many developers of face recognition algorithms adopted a biologically inspired approach in solving this problem (for a review, see Calder et al. [2001]), thus contributing both to understand human face processing and to build efficient face recognition technologies. Recent developments in the neurophysiology field inspired the development of a high-performance face recognition approach described in the present paper. The proposed approach is based on features that may be analogous to those extracted by the human visual system (HVS) from the visual scene. In particular, we evaluated the performance of a face recognition algorithm whose primary features were the magnitude of Cartesian or radial and angular components of images of faces. We show that the proposed approach is comparable to the state-of-the-art algorithms tested on the same databases, to the best of our knowledge. Possible implications for human face recognition are also discussed.

___

[1] Authors' addresses: Dept. of Computer Science – IME, University of São Paulo, São Paulo – SP 05508-900, Brazil, `[zana, roberto.cesar]@ime.usp.br`.

## 2. BACKGROUND AND PREVIOUS WORKS
### 2.1 Spatial Analysis in Polar Coordinates

Most of the current face recognition algorithms are based on feature extraction from a Cartesian perspective, typical to most analog and digital imaging systems. The primate visual system, on the other hand, is known to process visual stimuli logarithmically. For example, biological evidences indicate that the retinal image is retinothopically mapped onto area V1 of the visual cortex in a log-polar manner, i.e. image representation in the cortex is negatively correlated with retinal cell eccentricity [Schwartz 1977]. This property led to a formulation of a spatial log-polar transformation [Schwartz 1980] in which a Cartesian image was re-sampled as a logarithmic function of the distance from the center. This transformation was explored by several feature detection [Grove and Fisher 1996; Lim et al. 1997; Gomes and Fisher 2003] and face detection [Hotta et al. 1998; Jurie 1999; Chien and Choi, 2000] investigators. The log-polar transformation was also used in face recognition systems [Tistarelli and Grosso, 1998; Minut et al. 2000; Escobar and Ruiz-de-Solar 2002; Smeraldi and Bigun, 2002]. One of the disadvantages of this feature extraction method is the rough representation of peripheral regions. The HVS compensates this effect by eye saccades, moving the fovea from one point to the other in the scene. Similar approach was adopted by the face recognition systems of Tistarelli and Gross [1998] and Smeraldi and Bigun [2002]. Escobar and Ruiz-del-Solar [2002] applied the log-polar transformation at a single location, but the subsequent step (Gabor jets filtering) required manual localization of 16 fiducial points in the face image.

Although the previous algorithms that use log-polar transformation simulate well the retinal sampling resolution, they are usually followed by local analysis and do not provide any information about global patterns. An alternative representation of an image in the polar frequency domain is the two-dimensional Fourier-Bessel Transform [Bowman 1958; Rosental et al. 1982]. This transform found several applications in analyzing patterns in a circular domain [Zwick and Zeitler 1973; Guan, et al. 2001; Fox et al. 2003], but was seldom exploited for image recognition. One of such rare examples is the work of Cabrera et al. [1992], who applied the FBT to create descriptors of contour segments from images.

The paper is organized as follows: in the next section, we briefly introduce the reader to the primary spatial processing by the HVS and to the main face recognition algorithms tested on the face databases used in our experiments. We describe in section 3 the Discrete Fourier Transform (DFT) and the Fourier-Bessel Transform (FBT) for face image analysis. The proposed face recognition algorithms are introduced in section 4. We present the experimental results on two widely used face databases in section 5. In the last section we discuss the results and ongoing work for possible future improvements.

2.2 Human Spatial Processing

In the early stages of the human visual processing, specialized neurons act as filters for the visual image. Such neurons are tuned to specific spatial frequencies and locations in the visual field. The activity of cortical simple cells is well described by linear models [De Valois and De Valois 1990; Itti et al. 2000]. However, the local information provided in the earlier cortical stages must be further processed to extract global shape information. For instance, neurons in the inferotemporal cortex tuned to complex shapes such as faces have been reported [Perret et al. 1982]. Moreover, the response of these face-specific neurons was correlated with two-dimensional patterns, but not with three-dimensional shapes or any internal configural relations of the pattern [Young and Yamane, 1992].

Since the pioneer work of Kelly [1960], who suggested probing the HVS with symmetrically circular stimuli, many investigators of global shape processing in the early visual stages used circular shapes and found evidence of polar visual form processing. Electrophysiological experiments showed that cells in the LGN, V1, V2 and V4 cerebral areas in monkeys are specifically sensitive to Cartesian, polar and hyperbolic stimuli [Gallant et al. 1993, 1996; Mahon and De Valois 2001]. Moreover, psychophysical measurements of Glass dot patterns detection thresholds as a function of the stimulated area showed global pooling of orientation information in the detection of angular and radial dot patterns [Wilson and Wilkinson 1998]. Thus, it is evident that information regarding the global polar content of images is effectively extracted by and available to the HVS. Examples of the stimuli used by these investigators are presented in Figure 1.

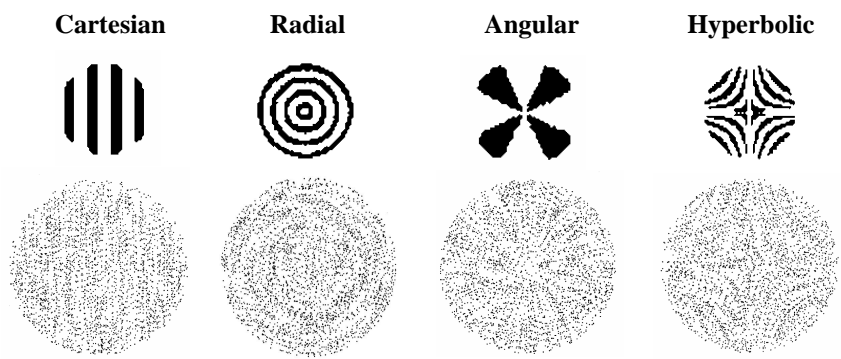

Fig. 1: Examples of stimuli defined in four different coordinate systems. Upper row: Stimuli used to study the response specificity of cells in the V4 cerebral area of monkeys [Gallant 1996]. Lower row: Glass patterns used for measuring of the detectability of dot patterns [Wilson and Wilkinson 1998].

Does polar form analysis help in face recognition tasks? To answer this question we need to evaluate the amount of information about the subject identity carried by polar components. Furthermore, we have to verify that the information is actually used by the HVS to recognize faces. In the current paper, we describe an attempt to answer the former question by developing a new efficient face recognition method.

2.3 Previous Face Recognition Algorithms

One of the fastest and most used face recognitions algorithms is the eigenfaces scheme [Turk and Pentland, 1991]. The first eigenvectors (principal components) obtained from a set of training images form the basic representation in that approach. Eigenfaces, the projection of the images onto these eigenvectors, are used as classification features. At the testing stage, unlabeled (probe) images are projected onto the eigenvector basis and compared to the learned images in the new face space. PCA was used to model facial expression recognition [Calder et al. 2001] and several known face effects, like distinctiveness, caricature and "other-race" (see [Calder et al. 2001] for a detailed review). There is also an interesting interpretation of the eigenfaces algorithm related to neurons specifically tuned to face images and to biological autoassociative memory models [Abdi, 1988; O´Toole et al. 1995]. The classic eigenfaces method is a good example of a global (holistic) algorithm, in contrast to feature-based (local) algorithms, where the image is encoded as a whole. A major drawback of the method is its requirement of a precise spatial normalization of the face parts, such that the eyes, mouth, etc., are scaled and registered at the same spatial coordinates [Craw and Cameron, 1991].

Also taking a holistic approach, Hafed and Levine [2001] explored the Discrete Cosine Transform (DCT) as a means of feature extraction for face classification. Analogous to the Fourier transform, the DCT maps an image from the spatial to the frequency domain, extracting the frequency components. However, it differs from the DFT in using only real coefficients. The biological appeal of the DCT features lies in the spatial to (Cartesian) frequency domain transformation that is believed to occur in the HVS [Campbell and Robson, 1968].

An algorithm developed in the University of Southern California proposed the Gabor wavelets image transform as mean of feature extraction [Lades et al. 1993; Wiskott et al. 1997]. This transform can be seen as analogous to the multiscale multiorientation spatial analysis observed in the V1 cortical area [Hubel and Wiesel, 1968; Movshon et al. 1978; De Valois et al. 1982]. In contrast to the global analysis of the DCT, the Gabor wavelets features are extracted at fiducial points, corresponding to anatomically identifiable nodes in a geometric model (graph) of the face. Probe images are projected onto the Gabor jets and compared, through elastic graph matching, to the learned images based on both the extracted coefficients at the corresponding locations and distances between the formed graphs. This method requires manual definition of the grid structure, but a fully automatic version was published recently [Arca et al. 2003].

Etemad and Chellappa [1997] used a hybrid approach in which images were initially represented by facial features in the spatial domain as well as by the wavelet transform. In a second stage, new features were extracted through Linear Discriminant Analysis (LDA). In contrast to PCA, the features are projected in LDA such that the (between-subjects variation)/(within-subjects variation) ratio is maximized. Moghaddam et al. [2000] proposed a Bayesian generalization of the LDA method where, instead of matching images through a Euclidean-based similarity measurement, the a posteriori probability of the difference between images is estimated considering the within- and between-subject variation in the training set. All computations were done in a PCA dimensionality reduced space.

Lawrence et al. [1997] proposed a different hybrid algorithm that combined local image sampling, a self-organizing map (SOM) neural network, and a convolutional neural network (CN). The SOM provides a quantization of the image samples into a topological space, where inputs that are nearby in the original space preserve proximity in the output space, while the CN network extracts successively larger features in a hierarchical set of layers. The SOM topological mapping can be associated to the retinotopic maps found in the visual cortex [Obermayer et al. 1991], while the hierarchical layers of the CN networks are analogous to the multiscale spatial sampling of the HVS [De Valois and De Valois, 1990].

As we expect our algorithm to be comparable not only from a biological perspective, but also from the practical point-of-view, non-biologically motivated algorithms are also cited here. One of these algorithms was designed by Samaria and Harter [1994], being based on hidden Markov models (HMMs) of a spatial top-down sampling. Samaria [1994] also published an extended version with pseudo two-dimensional HMMs. Another algorithm of interest is the ARENA [Sim et al. 2000], based on dimensionality reduction by simply averaging non-overlapping regions in the images. Then, it matches training and probe images basically by counting the number of components that differ in value (the $L_p^*$ dissimilarity measurement).

## 3. IMAGE TRANSFORMS

### 3.1 Discrete Fourier Transform

Spatial frequency analysis in Cartesian coordinates is traditionally done by applying the two-dimensional DFT and was already used for face recognition. For example, Akamatsu et al. [1991] applied the eigenfaces method to the magnitude of the Fourier spectrum as a mean of reducing variability due to changes in head orientation and shifting. The DFT is a well known analysis tool and will be briefly described. The equation to compute the DFT on an $M$ x $N$ size image is

$$F(u,v) = \frac{1}{\sqrt{MN}} \sum_{x=0}^{M-1} \sum_{y=0}^{N-1} f(x,y) e^{-j2\pi(\frac{ux}{M}+\frac{vy}{N})}$$

where $u$ and $v$ are the coordinates in the Fourier domain and $x$ and $y$ the coordinates in the space domain. The DFT deals with complex numbers that represent the magnitude and phase of the sine and cosine waves in the Fourier formula. However, in the present study only the magnitude $|F(u,\mathbf{n})|$ was considered.

### 3.2 Fourier-Bessel Transform

Let $f(x,y)$ be the face image. The FBT analysis starts by converting the image pixels description from Cartesian $(x,y)$ to polar $(r,\mathbf{q})$ coordinates. Let $(x_o, y_0)$ be the origin of the Cartesian image. The polar coordinates necessary to obtain the new image representation $f(r,\mathbf{q})$ are defined as

$$q = \tan^{-1}\left(\frac{y - y_0}{x - x_o}\right) \qquad (1)$$

and

$$r = \sqrt{(x - x_0)^2 + (y - y_0)^2} \qquad (2)$$

For square images, the considered maximum radius was the distance from the center of the image to one of the corners. Points outside the original image were discarded. Radial resolution was fixed at one pixel width, but the angular resolution could be varied by increasing or reducing of the number of sampled radii. The intensity of each point of the $f(r,q)$ function was determined by bilinear interpolation, combining the values of the four closest pixels weighted by their relative proximity to the reference point [Pratt, 1991]. Although this is a linear-polar transformation, in contrast to the aforementioned log-polar transformation, the central area of the Cartesian image is more densely sampled than the periphery in practice, especially at the high angular resolution used in the current study, due to the limited resolution and discrete nature of digital images,.

The FBT notation and definitions follow Bowman [1958] and Spanier and Oldham [1987]. The $f(r,q)$ function is than represented by the Fourier-Bessel series. The Bessel function of the first kind of order $n$ is defined by

$$J_{n(x)} = \left(\frac{x}{2}\right)^n \sum_{k=0}^{\infty} \frac{\left(\frac{-x^2}{4}\right)^k}{k!\Gamma(n+k+1)} \qquad (3)$$

where $\Gamma(x)$ is the gamma function. Figure 2 shows a graphical representation of three Bessel functions.

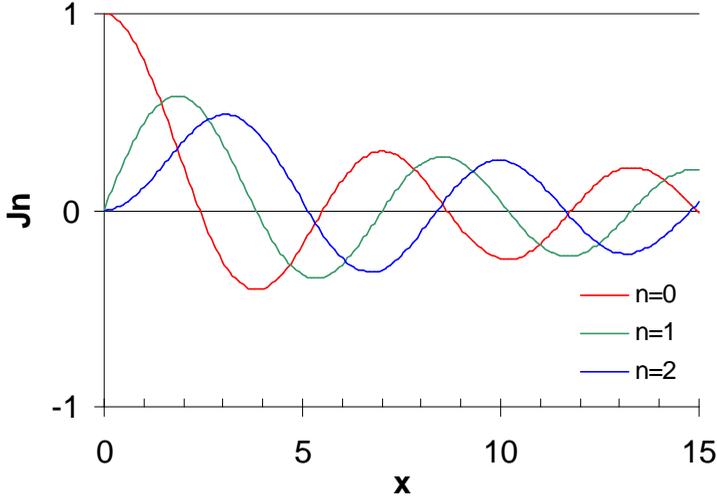

Fig. 2: A plot of the Bessel functions of the first kind of order 0, 1, and 2.

The two-dimensional Fourier-Bessel series is defined as

$$f(r,\theta) = \sum_{i=1}^{\infty}\sum_{n=0}^{\infty} A_{n,i} J_n(\alpha_{n,i} r)\cos(n\theta) + \sum_{i=1}^{\infty}\sum_{n=0}^{\infty} B_{n,i} J_n(\alpha_{n,i} r)\sin(n\theta) \quad (4)$$

where $f(R,\theta) = 0$ and $0 \leq r \leq R$. $\alpha_i$ is the $i^{th}$ root of the $J_n$ function, i.e. the zero crossing value satisfying $J_n(\alpha_{n,i}) = 0$. $R$ is the radial distance to the edge of the image. The orthogonal coefficients $A_{n,i}$ and $B_{n,i}$ are given by

$$A_{0,i} = \frac{1}{\pi R^2 J^2_1(\alpha_{n,i})} \int_{\theta=0}^{\theta=2\pi}\int_{r=0}^{r=R} f(r,\theta) r J_n(\frac{\alpha_{n,i}}{R}r) dr d\theta \quad (5)$$

if $B_{0,i} = 0$ and $n = 0$.

$$\begin{bmatrix} A_{n,i} \\ B_{n,i} \end{bmatrix} = \frac{2}{\pi R^2 J^2_{n+1}(\alpha_{n,i})} \int_{\theta=0}^{\theta=2\pi}\int_{r=0}^{r=R} f(r,\theta) r J_n(\frac{\alpha_{n,i}}{R}r) \begin{bmatrix} \cos(n\theta) \\ \sin(n\theta) \end{bmatrix} dr d\theta \quad (6)$$

if $n > 0$.

Some FBT examples are presented in Figure 3. In the original images, a sine function is plotted in radial (8 cycles per image) or angular (4 cycles per 360°) coordinates, and the third (right) image is the average of both images. Firstly, we transformed the coordinates of the pixels from Cartesian to polar using Eq. 1 and 2, with a 0.5° angular resolution (i.e. 720 radii). We calculated the coefficients A and B using Eq. 5 and 6, limiting $n$ to 30 Bessel order and $i$ to 30 Bessel root. These coefficients represent the magnitude of a spatial variation in polar coordinates [Fox, 2000; Guan et al. 2001]. For example, the 0th order, 8th root and 4th order, 1st root coefficients represent the magnitude of a pure 8 radial cycles or 4 angular cycles pattern, respectively. Similarly, the 4th order, 8th root coefficient represents a pattern which is the product of these two forms. The A and B coefficients represent the same patterns but in opposite phase. We plotted the modulus of A and B in Figure 3 for better visualization's sake. In this spectrum plot, the Bessel order columns represent the relative magnitude of the angular frequency while the Bessel root rows represent the radial frequency. It can be noted that the FBT of the images correctly indicates the principal radial and angular components. The inverse transform images can be calculated from the coefficient matrices (not showed in the figure).

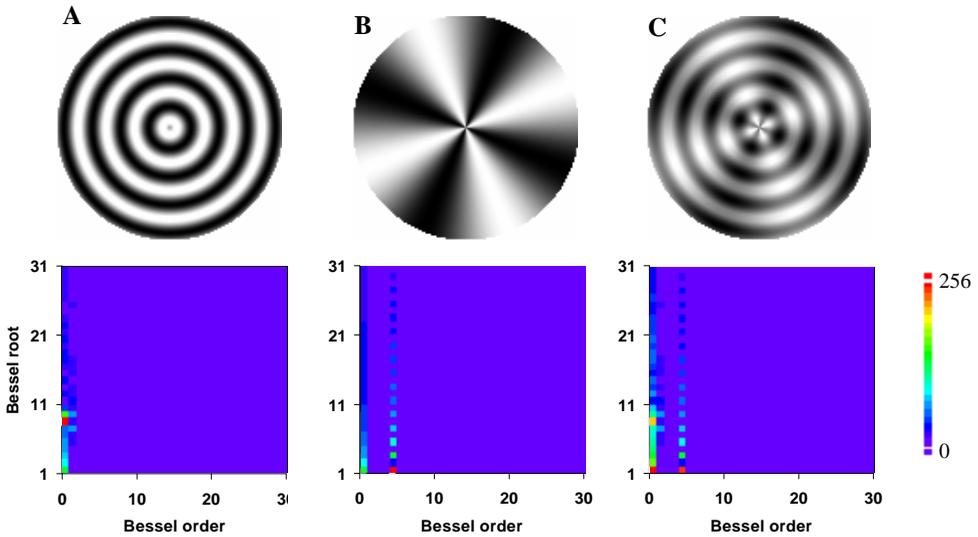

Fig. 3: Examples of FBT of (A) an 8 radial cycles image, (B) a 4 angular cycles, (C) and an image of the average of these images. Image size was 131 x 131 pixels. Angular resolution was 0.5°. The modulus of the FBT coefficients is presented in colored levels (red indicates the highest value).

## 4. ALGORITHMS

### 4.1 Feature Extraction

Dataset handling, training and tests were done with the Matlab PRTools toolbox [Duin, 2000]. After a pre-processing stage (detailed below), images were transformed by a FBT up to the 30th Bessel order and 3rd root with angular resolution of 0.5°, thus leading to 186 coefficients. DFT analysis extracted a total of 1200 magnitude coefficients, corresponding to up to 19.5 cycles per image. The number of DFT coefficients was determined by the optimal performance in preliminary tests on the ORL database.

Figure 4 shows an example of the FBT spectrum (extended to the 30th Bessel root) of a face image from the ORL database [Samaria and Harter, 1994]. The spectrum is of relatively complexity, but with low frequency components predominance. From the blurred aspect of the face inverse transform using all the coefficients it can be noted that the FBT represents well low to middle range frequencies, but not high frequencies. This is a consequence of using a limited number of Bessel orders and roots. Increasing the number of coefficients would widen the sampled frequency range, but also the computation time. However, although the original image cannot be exactly reconstructed (Fig. 4), only a limited frequency range is necessary to achieve good recognition rates, as shown below. In the rest of this Section, we will refer to the Fourier or Fourier-Bessel transformed images as just images.

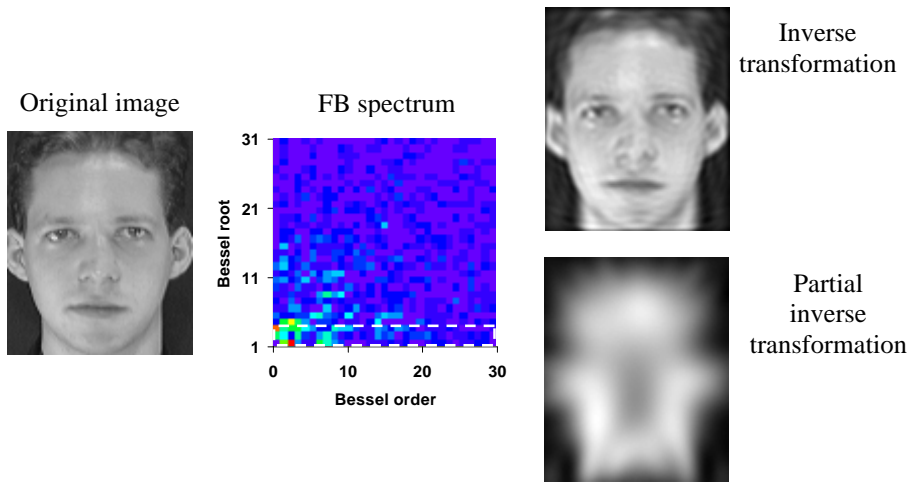

Fig. 4: FBT of a face image from the ORL database. Inverse transforms were derived from the full FB spectrum (upper image) or only up to the 3-rd Bessel root (lower image). The dashed line limits the FB coefficients used for the partial inverse transformation. The lowest coefficient in the spectrum image was set to minimum in order to improve the visibility of the other coefficients.

## 4.2 Dissimilarity Space

We built a dissimilarity space $D(t,t)$ defined as the Euclidean distance between all training images, where $t$ is the training image. In this space, each object is represented by its dissimilarity to all objects (Table I). This approach is based on the assumption that the dissimilarities of similar objects to "other ones" is about the same. Dissimilarity space for pattern recognition was first formulated by Duin et al. [1997] and successfully used with the Pseudo-Fisher Linear Discriminant (PFLD) on the ORL face database [Pekalska and Duin, 2000]. However, while Pekalska and Duin [2000] represented the 256 x 256 (65536) pixel intensity values in the dissimilarity space, we used 186 coefficients of the FBT. Among other advantages of this representation space, by fixing the number of features (or dimensions) to the number of objects, it avoids a phenomenon known as "peaking" [Raudys and Pikelis, 1980], where recognition performance is degraded as a consequence of small number of training samples as compared to the number of features.

Table I: An example of a dissimilarity space $D(t,t)$. The value in each cell indicates the Euclidean distance between the corresponding images. In this space, each image is represented by its dissimilarity to all images, i.e. by the values in the corresponding column.

|         | Image 1 | Image 2 | Image 3 |
|---------|---------|---------|---------|
| Image 1 | 0.0     | 0.8     | 0.3     |
| Image 2 | 0.8     | 0.0     | 0.5     |
| Image 3 | 0.3     | 0.5     | 0.0     |

## 4.3 Classifier and Testing

Test images were classified based on a PFLD using a two-class approach. A Fisher Linear Discriminant (FLD) is obtained by maximizing the Fisher criterion [Fukunaga, 1990], i.e. the (between subjects variation)/(within subjects variation) ratio. Here we used a minimum-square error classifier implementation [Skurichina and Duin, 1996], which is known to be equivalent to the FLD for two-class problems [Fukunaga, 1990]. In these cases, after shifting the data such that it has zero mean, the FLD can be defined as

$$g(D(\mathbf{x},t)) = \left[ D(\mathbf{x},t) - \frac{1}{2}(\mathbf{m}_1 - \mathbf{m}_2) \right]^T \mathbf{S}^{-1}(\mathbf{m}_1 - \mathbf{m}_2) \qquad (7)$$

where $D(\mathbf{x},t)$ is a probe image, $\mathbf{S}$ is the pooled covariance matrix, and $\mathbf{m}_i$ stands for the mean of class $i=1,2$. $D(\mathbf{x},t)$ is classified as corresponding to class-1 if $g(D(\mathbf{x},t)) \geq 0$ and to class-2 otherwise. However, as the number of training objects and dimensions is the same in the dissimilarity space, the sample estimation of the covariance matrix $\mathbf{S}$ becomes singular, and the classifier cannot be built. One solution to the problem is to use a pseudo inverse and augmented vectors [Skurichina and Duin, 1996]. Thus, Eq. 1 is replaced by

$$g(D(\mathbf{x},t)) = (D(\mathbf{x},t),1)(D(t,t),I)^{(-1)} \qquad (8)$$

where $(D(\mathbf{x},t),1)$ is the augmented vector to be classified and $(D(t,t),I)$ is the augmented training set. The inverse $(D(t,t),I)^{(-1)}$ is the Moore-Penrose Pseudo Inverse which gives the minimum norm solution. The pseudo-inverse relies on the singular value decomposition of the matrix $(D(t,t),I)$ and it becomes the inverse of $(D(t,t),I)$ in the subspace spanned by the eigenvectors corresponding to the non-zero eigenvalues. The classifier is found in this subspace [Pekalska and Duin, 2000].

The current $L$-classes problem can be reduced and solved by the two-classes solution described above [Allwein et al. 2000; Duin 2000]. The training set was splitted into $L$ pairs of subsets, each pair consisting of one subset with images from a single subject and a second subset formed from all the other images. A PFLD was built for each pair of subsets. A probe image was tested on all $L$ Discriminant functions, and assigned to the subject that gave the largest posterior probability.

### 4.4 Combining Classifiers

The FBT- and DFT-based algorithms classified the test images independently. Although both orthogonal bases can lead to a perceptually good reconstruction, they encode different type of information (Cartesian vs. polar frequency content). To evaluate the correlation between the classification errors made by the two algorithms, the soft-output of the classifiers were combined by a maximum rule before labeling the images, i.e. the final assignment was determined by the classifier that gave the highest normalized output [Kittler et al. 1998].

Figure 5 present a schematic description of the methods presented in this paper, i.e. (a) FBT-only, (b) DFT-only, and (c) FBT+DFT through maximum rule.

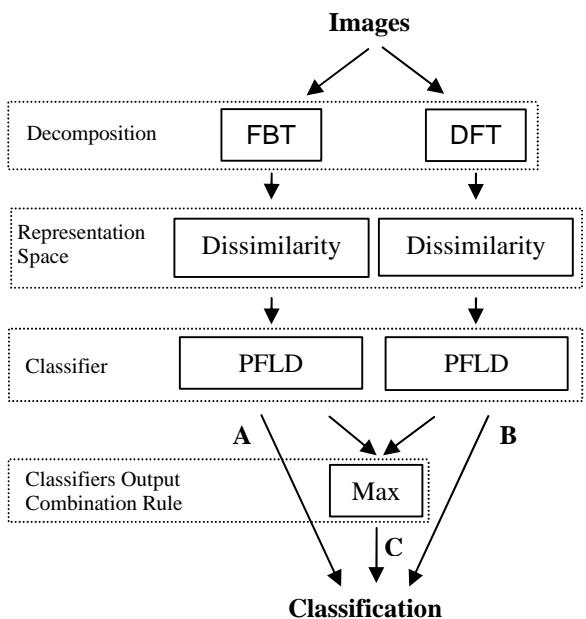

Fig. 5: Block diagram of the three face recognition algorithms: (A) FBT-only, (B) DFT-only, and (C) FBT+DFT through maximum rule.

## 5. EXPERIMENTAL RESULTS

We chose to test our algorithm on two distinct databases: the ORL [Samaria and Harter 1994] and the FERET [Phillips et al. 1998]. The former is a small set with only 40 subjects. However, 10 sample images from each subject, taken with small variations (see details below), are available. The FERET set includes images from thousands of subjects, but only a few samples from each. Both datasets were used as test platforms for many algorithms, making ranking comparisons easier.

### 5.1 The ORL database
#### 5.1.1 Database and performance evaluation

The ORL face database consists of 400 images collected from 40 people. Most of the subjects had 20-35 years. The face images were 92 x 112 pixels with 8-bit gray levels. They included variations in facial expression, luminance, scale and viewing angle and were shot at different time. Limited side movement and tilt of the head were tolerated. Some subjects are captured with and without glasses. These characteristics introduce difficulties to correct recognition and make the database particularly interesting. All images were manually cropped and rescaled to the final resolution by the authors of the ORL database [Samaria and Harter 1994] and no further graphical pre-processing was applied here. Figure 6 shows an example of an image set of one subject. Error rate was calculated as the percentage of the misclassified images and each test was repeated 10 times.

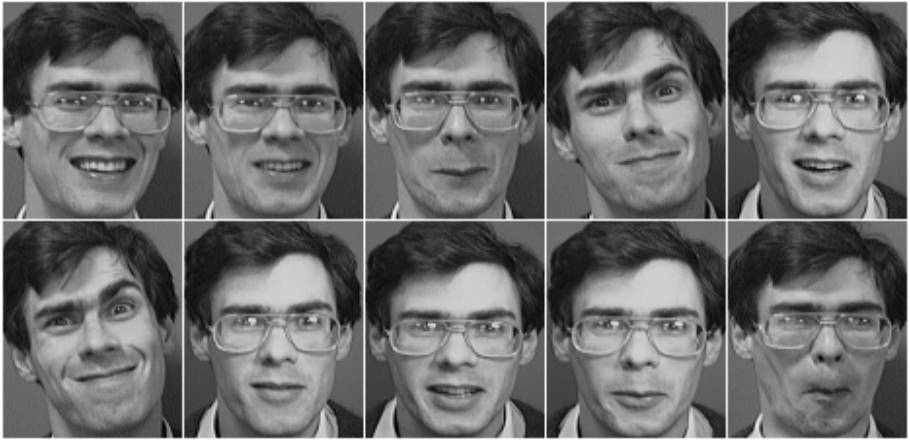

Fig. 6. Images from one subject of the ORL database. There are variations in illumination, head position, facial expressions and occlusion.

### 5.1.2 General performance and learning rate

Figure 7 shows the final error rates of the three algorithms, and a reproduction of results from previously published papers obtained with different models, but where the same database, sampling and testing methodology were used [Samaria and Harter 1994; Samaria 1994; Lawrence et al. 1997; Sim et al. 1999; Pekalska and Duin, 2000; Hafed and Levine 2001]. With five images per subject, the FBT achieved a final performance of 3.8%, while the DFT had an error rate of 1.25%. However, when the FBT algorithm and the DFT where combined, error rate was strongly reduced to 0.2%. This means that on average, of every 500 test images, only one was misclassified, being an indication that the classification errors made by the FBT and DFT are successfully complementary. The performance of the FBT is equal to that achieved by the SOM-CN algorithm, but lower than that of the most successful methods (ARENA and RAW-PFLD). However, the DFT and the FBT+DFT algorithms outperformed all the others.

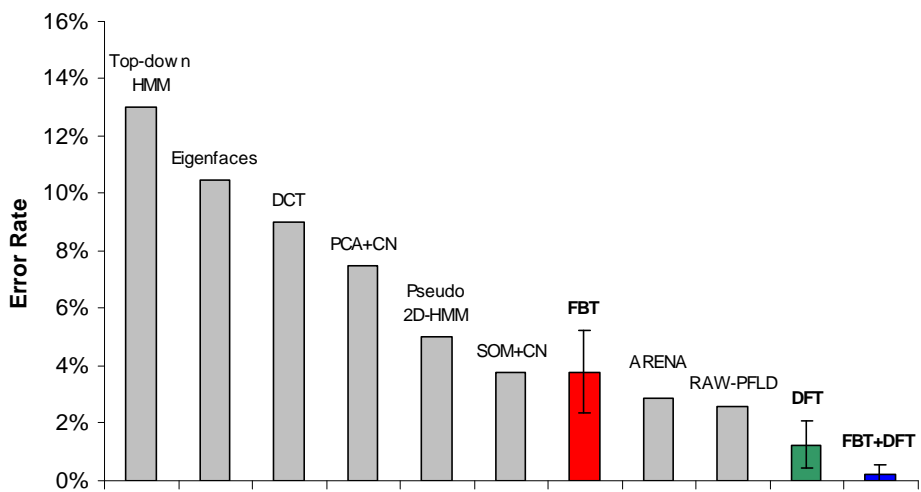

Fig. 7: Error rate of FBT, DFT and FBT+DFT algorithms. Training size was 5 images per subject. Error bars indicate ±1 standard error of the mean. Results of previous algorithms are from Samaria and Harter [1994] (Top-down HMM); Samaria [1994] (Pseudo 2d-HMM); Lawrence et al. [1997] (Eigenfaces, PCA+CN, SOM+CN); Hafed and Levine [2001] (DCT); Sim et al. [1999] (ARENA); Pekalska and Duin [2000] (RAW-PFLD).

Figure 8 shows the error rate as a function of the number of images from each subject, i.e. the learning rate. For all cases, the error rate decreases as the number of training samples increase, as expected, since the increasing number of different samples from each subject increases the chances of correct recognition when presented with a new image. The performance of both FBT and DFT algorithms, for training sets of less than five images per subject, was superior to all previous algorithms tested in the same modality. However, the combined FBT+DFT classifier achieved again the lowest error rate.

Lai et al. [2001] developed a face recognition system based on a spectroface representation, i.e. a wavelet transform and global Fourier invariant features, and presented results for the ORL database [Lai et al. 2001]. Their system achieved a 5.36% error rate when the training set included three images per subject. This performance falls between those of the FBT and DFT. However, their results are optimistically biased, as the three training images were selected so as to form the best face representation.

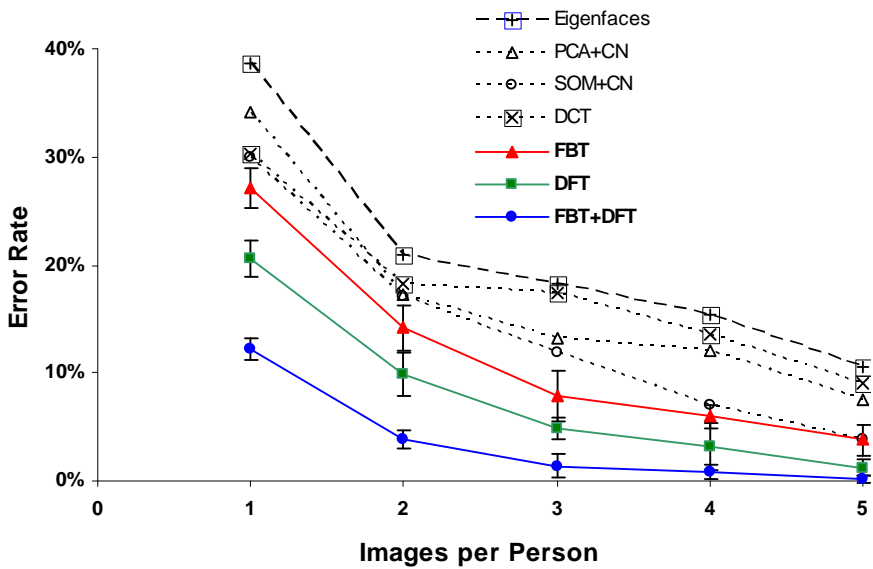

Fig. 8: Error rate as a function of the number of images learned from each subject. Error bars indicate ±1 standard error of the mean. . Results of previous algorithms are from Lawrence et al. [1997] (Eigenfaces, PCA+CN, SOM+CN); Hafed and Levine [2001] (DCT).

### 5.1.3 Dataset size effect

The performance of face recognition algorithms usually degrades as more subjects are added to the database, due to the increasing probability of the presence of subjects with similar attributes. This effect was confirmed with the models presented here (Fig. 9). A comparable test on the same database was available only for the the SOM+CN algorithm [Lawrence et al. 1997]. Again, the performance of the FBT, DFT and the combined classifier were found to be superior.

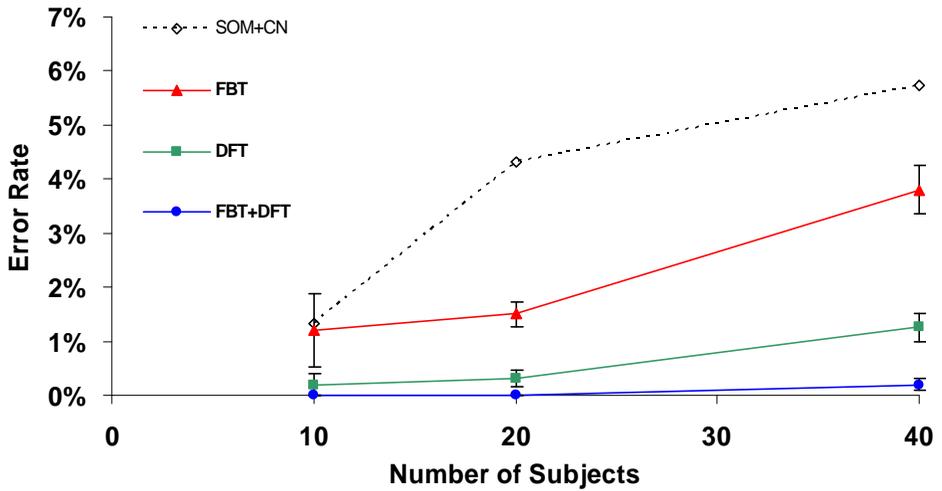

Fig. 9: Error rate as a function of the number of subjects in the database. Training set size was 5 images per person in all cases. Error bars indicate ±1 standard error of the mean. Results for the SOM+CN algorithm are from Lawrence et al., 1997.

### 5.1.4 Performance of individual features

We have also analyzed the performance of the FBT and DFT individual features. For this test, the number of extracted coefficients was extended. The training and test sets were constituted of five images per subject, each image represented by a single feature. Recognition was based on a simple nearest neighbor criterion, were a test image is assigned to the subject of the closest training image. Figure 10 shows the error rates obtained by using each feature. The highest performance for the FBT was obtained by coefficients in approximately the 0-20 Bessel order range coupled to Bessel roots in the 1-6 range. This result indicates that angular components of low-to-medium frequencies coupled to low frequency radial components are the most informative with respect to the subject identity. The observation that the necessary features for good recognition performance are insufficient to reconstruct the original image (Fig. 4) agrees with previous studies [Turk and Pentland, 1991; Hafed and Levine, 2001] and does not violate any biological principle.

The DFT features presented a somewhat different behavior. The best performing features were distributed at all frequencies and orientations, although more concentrated in the low frequency range and horizontal orientation (y axis in the plot). This result may explain why the DFT algorithm requires such an extensive number of features, as compared to the FBT, to achieve good recognition levels.

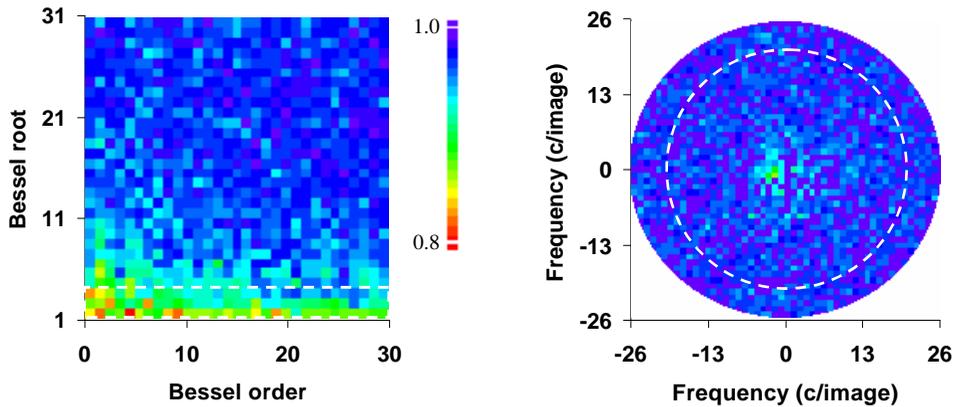

Fig. 10: Error rate, in colored levels, of individual features extracted by FBT (left panel) or (right panel) DFT. The DFT is presented in a conventional polar plot, where distance from the center represents frequency and angle represents orientation. Training size was 5 images per subject. Each pixel represents the average of 10 samples. The dashed white lines mark the spectral regions that were used in the FBT and DFT algorithms.

## 5.2 The FERET database
### 5.2.1 Database and performance evaluation

The FERET database was collected as part of the Face Recognition Technology program to support algorithm development and evaluation. The main advantages of this database are the large number of individuals and rigid testing protocols that allow precise performance comparisons between different algorithms. Basically, a "gallery" set of one frontal view image from 1196 subjects is used to train the algorithm and a different dataset is used as probe. All images are gray-scale 256 x 384 pixels size. For the current study, we used the largest probe set, termed "FB", which is constituted of a single image from 1195 subjects. The probe set images were taken from the same subjects in the gallery set, after an interval of a few seconds, but with a different facial expression.

The original images were normalized using the eyes ground-truth information supplied with the database. Images were translated, rotated, and scaled so that the eyes were registered at specific pixels. Next, the images were cropped to 118 x 140 pixels size and a mask was applied to remove most of the hair and background. No histogram equalization or luminance normalization was performed, once these operations showed no performance improvement in our tests.

The performance of the algorithms was evaluated by identification and verification tests according to the FERET protocol [Phillips et al. 1998]. In the identification test, the algorithm ranks the gallery images according to their similarity to the probe images. A good algorithm would give a low ranking score to the correct matching images. For example, if the correct match of a probe image is among the five most similar gallery images, the matching is ranked five. Results are plotted as the proportion of correct identification as a function of the rank. Thus, the performance of an algorithm that matches 80 out of 100 probe images with ranking of five or less will be 80/10=0.8 at rank 5.

The verification test is based on an open-universe model. Given a gallery image $g$ and a probe image $p$, the algorithm verifies the claim that both were taken from the same subject, i.e. that $p \approx g$. The verification probability $P_V$ is the probability of the algorithm accepting the claim when it is true, and the false-alarm rate $P_F$ is the probability of incorrectly accepting a false claim. The algorithm decision depends on a "confidence" ("posterior probability") score $s_i(k)$ given for each match and on a threshold $c$. Thus, a claim is confirmed if $s_i(k) \leq c$ and rejected otherwise. A plot of all the combinations of $P_V$ and $P_F$ as a function of $c$ is known as a receiver operating characteristic (ROC). For each probe image, we computed the posterior probabilities for all the subjects in the gallery set and applied a threshold function. The results were splitted to a subset that consisted of all matches where $p \approx g$, and a second subset with all matches where $p \neq g$. $P_V$ and $P_F$ were calculated as the number of confirmations divided by the number of matches in each subset, respectively. This procedure was repeated for 100 equally spaced threshold levels.

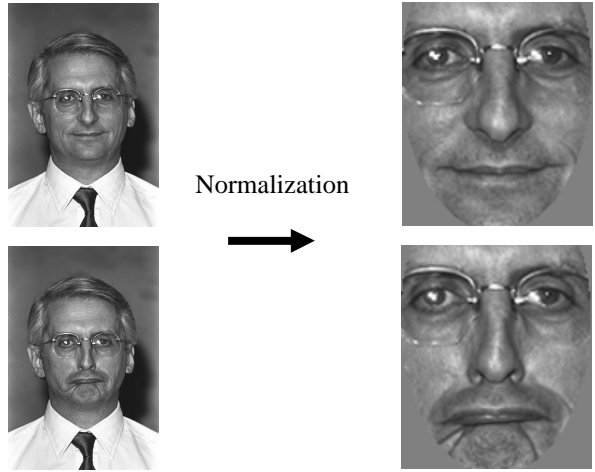

Fig. 11: Example of a normalization of two images taken from the same subject. The original images were scaled, aligned, centralized, cropped, and masked.

### 5.2.2 Face identification and verification

Figure 12A shows the correct identification rate as a function of the rank for the proposed methods. In the same plot we present results for the main previous algorithms, as published in Phillips et al. [1998]. The performance of the FBT algorithm was slightly superior to the basic PCA method, but lower than the state-of-the-art algorithms. In contrast to the results on the ORL database, here the DFT algorithm performed significantly worse than the FBT and no improvement was achieved by using the combined algorithm.

We also tested the proposed algorithms in a verification modality (12B). In this case the FBT algorithm outperformed the PCA and the PCA+Bayesian algorithms. It equaled the Gabor+EBGM method at 0.005 false acceptance probability level, but did not reach the PCA+LDA performance. An interesting performance indicator is the equal error rate, i. e. the point where incorrect rejection and false alarm rates are equal. Figure 13 shows the results for the FBT algorithm in detail, while Table II summarizes results from previous studies. A comparative analysis of the equal error rate shows an even better performance of the FBT method: the error rate was lower than the PCA and PCA+Bayesian algorithms, equal to the Gabor+EBGM, but higher that of the PCA+LDA method. Here again, the DFT and the combined algorithm did not improved the performance over the FBT algorithm.

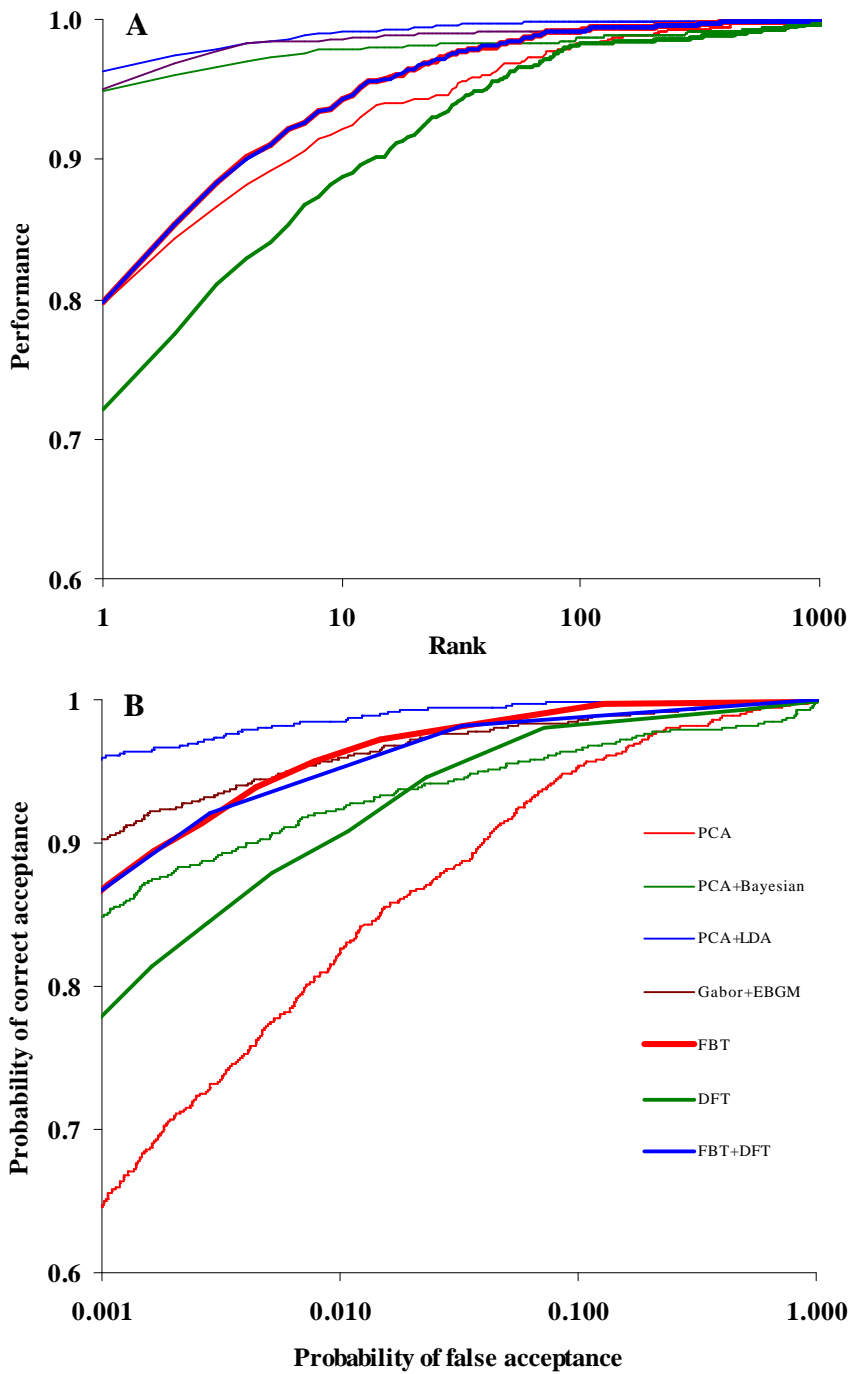

Fig. 12: Performance on a (A) ranked recognition and (B) verification tests. Results of previous algorithms are from Phillips et al. [1998], based on standard PCA (PCA), and works of Moghaddam et al. [2000](PCA+Bayesian), Etemad and Chellappa [1997](PCA+LDA), and Lades et al. [1993](Gabor+EBGM).

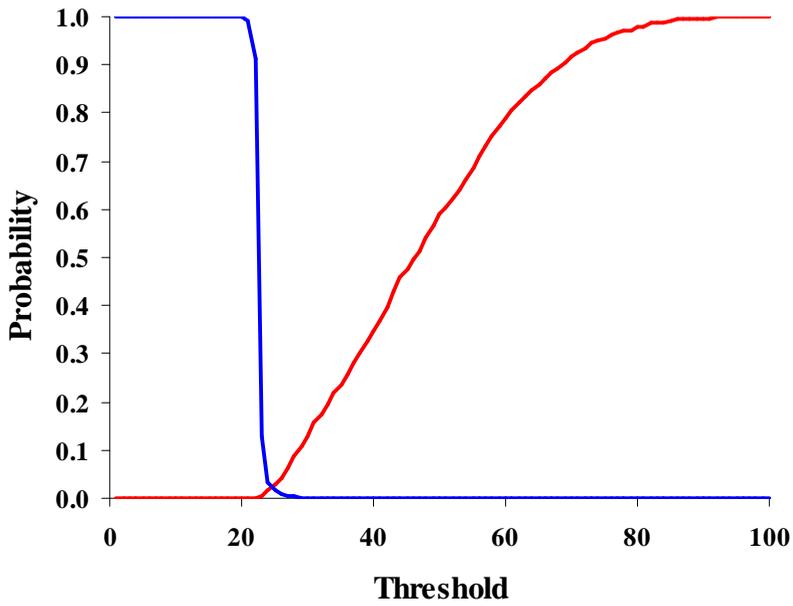

Fig. 13: Incorrect rejection (red line) and false alarm (blue line) rate as a function of the threshold of the FBT algorithm.

Table II: Equal error rate of the proposed and previous algorithms.

| Algorithm | Equal error rate |
|---|---|
| PCA | 7 |
| PCA+Bayesian | 4 |
| PCA+LDA | 1 |
| Gabor+EBGM | 2 |
| FBT | 2 |
| DFT | 4 |
| FBT+DFT | 3 |

## 6. DISCUSSION

We introduced here a biologically-motivated novel combination of techniques for face recognition tasks. The new algorithm achieved considerable performance, although not surpassing that of state-of-the-art algorithms under all test conditions. We believe that it works due to the highly informative value of the polar frequency components, the conservation of proximity relations in the dissimilarity space, and by the efficient class separation provided by the Linear Discriminant.

### 6.1 Polar Frequency Domain

Although the relation of the present algorithm to human face recognition was not directly evaluated here, a few associations can be done. As discussed in the Introduction, there is clear evidence that the HVS performs a Cartesian local analysis of the visual scene, but also pools this information to extract global radial and angular shapes. A model of such spatial processing was developed by Wilson and others [Wilson, 1991; Wilson et al. 1997; Wilson and Wilkinsom, 1998; Wilson et al. 2001]. Thus, in theory, the global polar frequency content of face images is available to the HVS, in addition to the local Cartesian frequency content.

In the current proposal we explored an analogous global polar pooling by applying a FBT. The analogy resides not in the way the Cartesian information is pooled to extract global components, but in the coordinate definition of the fundamental global patterns. The face recognition tests showed that only as little as 186 components are required for good performance, demonstrating the compactness of the polar representation. When tested on the ORL database, the FBT algorithm had achieved good performance compared to previous algorithms, but had the best performance when combined with a DFT-based algorithm. These results indicate coding of complementary information regarding face identity by the two transformation types. However, when tested on the larger FERET database, no performance gain was observed. These results suggest that the DFT contributed to the FBT performance by compensating for image variations such as translation, considering that the ORL images were not normalized, in contrast to the FERET images. This question should be addressed more specifically by future research.

The specific polar components can also be related to human performance. It was shown that the most informative individual features were found in the low-to-medium angular frequencies coupled to low radial frequencies, and that the FBT algorithm achieved good recognition performance when it was based on a wide range of angular frequencies coupled to low radial frequencies. We can not make a direct comparison of this finding with previous studies, since only Cartesian frequencies were considered in the latter case. It was evident, however, that the DFT-based algorithm required a wide range of frequency sampling in order to achieve good results, in agreement with the findings that both low and high spatial frequencies are important for face recognition algorithms [Harmon, 1973; Sergent, 1986; Nastar, et al. 1997]. Human face recognition, on the other hand, is tuned to a limited band of frequencies, between 10 and 20 cycles per face, according to psychophysical studies [Tieger and Ganz, 1979; Costen et al. 1996; Nasanen, 1999]. Other studies concluded that gender classification and identification can be based on solely low- or high-frequency components, respectively [Sergent, 1986]. Thus, if polar frequency components participate in human face recognition, it is expected that a wide range of angular frequencies, along with low radial frequencies, would have the highest relative weight for face recognition.

6.2 Dissimilarity Representation Space

In the proposed algorithm, the classifier operates in non-domain-specific metric space whose coordinates are similarity relations. In this space, images are represented as points whose coordinates are defined by their similarities to the other images, not by their FB coefficients. The effect of the PFLD is basically to reduce the distances between points of the same subjects and to increase it for points from other subjects. Probe images are also mapped onto this dissimilarity space, projected on the PFLD components, and classified according to the nearest class object. The high performance achieved by this representation indicates that the "real-world" proximity relations between face images are preserved to a good extent in the constructed internal space. An interesting question to ask is "Do humans internally represent faces in a dissimilarity space"?

One strategy to answer this question is to measure the dissimilarity (or distance) between different shape objects by objective (like geometry or pose) and perceptual (like recognition response time) parameters (for detailed review of shape representation by humans, see [Edelman, 1999]). Comparison of the two measurements is usually done by a multidimensional scaling (MDS) analysis, which projects objects as points in a two-dimensional space where the distance between the points approximate the Euclidean distance between the original objects. For example, in one study 41 face images were sorted by degree of similarity according to geometrical (manual) measurements and by perceptual similarity [Rhodes 1988]. The two sorting method results were highly correlated, especially when the objective sorting used global features, such as age and weight of the persons in the images. Similar results were obtained in a neurophysiological study [Young and Yamane, 1992] in which monkeys were presented with face images. It was found that the MDS proximity maps obtained from the original images and from the response patterns of neurons in the inferotemporal cortex had similar patterns.

Evidence of dissimilarity representation by humans comes also from studies that did not involve face images. In one of the first application of this method, subjects were submitted to memory recall tests of the outline shapes of 15 of the US states [Shepard and Chipman, 1970]. In other studies subjects were submitted to recognition tests of closed contours representing the outline shape of objects [Shepard and Cermak, 1973; Cortes and Dyre, 1996], while objective measurements were based on Fourier descriptors. In all these cases, MDS analysis of shapes and subject responses revealed a similar pattern. Extending the studies to animal-like shapes led to the same conclusion [Edelman, 1995; Cutzu and Edelman, 1996, 1998]. The task of the subjects was to decide if two objects are similar and which of two pairs of objects are more similar to each other. Again, based on delay time measurements, MDS analysis showed that people related the objects according to the objective similarity relations. Control experiments with "scrambled" (non-sense) shapes did not replicate the results.

These results indicate that representing images in a dissimilarity space, as done in the proposed algorithm, can be analogous to human representation mechanisms. It is important to note that in several of these studies, the shape measurements that correlated with the human performance were taken from a global perspective (age, Fourier descriptors, etc), a further indication of the relevance of global image analysis approach adopted here for human face recognition.

## 6.3 Future Research

Ideally, the representation of a face would be robust to changes in its appearance, but still be able to distinguish it from faces of other subjects. The FBT-based algorithm was tested only on frontal view face images and relatively constant illumination, and there is no reason to believe that it will perform well under strong variation of this kind. From the computational point of view, we expect that a face-tracking algorithm [Feris et al. 2004], followed by image normalization, would improve the algorithm performance and make it independent of ground-truth information. Preliminary tests also indicate that image resolution can be strongly reduced with only a small effect on performance, but with a significant reduction in processing time.

Performance improvement can be achieved by normalization of the FBT coefficients, such that the face representation becomes invariant to translation, rotation, and scale

[Cabrera et al. 1992]. This is possible if the transform is obtained from the center of a face whose contour (or radius) is known. The coefficients ranking test results indicate that the algorithm may beneficial from an implementation of a feature selection step. Currently all the coefficients in a specific range of frequencies are used. An alternative would be to increase the frequency range, but select only the coefficients with the highest predictive value, based on the training set. Finally, as the Euclidean distance currently in use was found to perform poorly [Sim et al. 1999; Moghaddam et al. 2000], other alternative metrics like the Mahalanobis distance should be evaluated.

An important issue to be investigated regards local and multiscale features, which should be integrated in our future experiments by applying local FBT and exploring Gabor filters and wavelets on polar representation of faces. Finally, our ongoing interests include face recognition in video sequences and in different 3D poses.

ACKNOWLEDGMENTS

This research was supported by grants from CNPq (150459/2003-3 , 301290/00-8, 478384/2001-7, 300722/98-2) and FAPESP (No: 03/07519-0 and 99/12765-2). We are grateful to Paul Fox, Viviana Giampaoli, Bob Duin, Elzbieta Pekalska, and Jesus Mena-Chalco for their valuable comments on the mathematical aspects of the work. We are also grateful to the reviewers for the useful comments that helped us improving the paper.

REFERENCES

ALLWEIN, E.L., SCHAPIRE, R.E., AND SINGER, Y. 2000. Reducing multiclass to binary: a unifing approach for margin classifiers. *Journal of Machine Learning Research* 1: 113-141.
ARCA, A., CAMPADELLI, P., LANZAROTTI, R. 2003. A face recognition system based on local feature analysis. *Lecture Notes in Computer Science* 2688: 182-189.
ABDI, H. 1988. A generalized approach for connectionist auto-associative memories: interpretation, implication and illustration for race process. In J. Demongeot, T. Hervé, V. Rialle, and C. Roche (Eds.). *Artificial Intelligence and Cognitive Sciences*. Manchester University Press, Manchester, pp. 149-165.
AKAMATSU, H.F.S., SASAKI, T., AND SUENUGA, Y. 1991. A robust face identification scheme – KL expansion of an invariant feature space. In *SPIE Proceedings: Intelligent Robots and Computer Vision X: Algorithms and Techniques* 1607: 71-84.
BOWMAN, F. 1958. *Introduction to Bessel Functions*. Dover Publications, New York, NY.
CABRERA, J., FALCÓN, A., HERNÁNDEZ, F.M., AND MÊNDEZ, J. 1992. A systematic method for exploring contour segment descriptions. *Cybernetics and Systems* 23: 241-270.
CALDER, A.J., BURTON, A.M., MILLER, P., YOUNG, A.W., AND AKAMATSU, S. 2001. A principal component analysis of facial expressions. *Vision Research* 41: 1179-1208.
CAMPBELL, F.W., AND ROBSON, J.G. 1968. Application of Fourier analysis to the visibility of gratings. *Journal of Physiology* 197: 551-566.
CHIEN, S.I, AND CHOI, I. 2000. Face and Facial Landmarks Location Based on Log-Polar Mapping. *Lecture Notes in Computer Science* 1811: 379-386.
CORTES, J. M. AND DYRE, B. P. 1996. Perceptual similarity of shapes generated from Fourier Descriptors. *Journal of Experimental Psychology: Human Perception and Performance* 22: 133-143.
COSTEN, N.P., PARKER, D.M., AND CRAW, I. 1996. Effects of high-pass spatial filtering on face identification. *Perception and Psychophysics* 58: 602-612.
CRAW, I. AND CAMERON, P. 1991. Parametising images for recognition and reconstruction. *Proceedings of the British Machine Vision Conference*: 367-370.

CUTZU, F., AND EDELMAN, S. 1996. Faithful representation of similarities among three-dimensional shapes in human vision. *Proceedings of the National Academy of Science* 93: 12046-12050.
CUTZU F., AND EDELMAN S. 1998. Representation of object similarity in human vision: psychophysics and a computational model. *Vision Research* 38: 2229-2257.
DE VALOIS, R.L., AND DE VALOIS, K.K. 1990. *Spatial Vision*. Oxford University Press, New York, NY.
DE VALOIS, R.L., YUND, E.W., AND HELPER, N. 1982. The orientation and direction selectivity of cells in macaque visual cortex. *Vision Research* 22: 531-544.
DUIN, R.P.W. 2000. *PRTools_3, A Matlab Toolbox for Pattern Recognition*. Delft University of Technology.
DUIN, R.P.W., DE RIDDER, D., AND TAX, D.M.J. 1997. Experiments with a featureless approach to pattern recognition. *Pattern Recognition Letters* 18: 1159-1166.
EDELMAN, S. 1995. Representation of similarity in 3D object discrimination. *Neural Computation* 7: 407-422.
EDELMAN, S. 1999. *Representation and Recognition in Vision*. MIT Press, Cambridge.
ESCOBAR, M.J., AND RUIZ-DEL-SOLAR, J. 2002. Biologically-based face recognition using Gabor filters and log-polar images. *Proceedings of the International Joint Conference on Neural Networks 2: 1143-1147*.
ETEMAD K, CHELLAPPA R. 1997. Discriminant analysis for recognition of human face images. *Journal of the Optical Society of America A-Optics Image Science and Vision* 14: 1724-1733.
FERIS, R.S., KRUEGER, V., AND CESAR-JR, R.M. 2004. A wavelet subspace method for real-time face tracking. *Journal of Real-Time Imaging*, in press.
FOX, P.D. 2000. Computation of linear ultrasound fields using 2D Fourier-Bessel series. *Proceedings of the 25th International Acoustical Imaging Symposium*: 19-22.
FOX, P.D., CHENG, J., AND LU, J. 2003. Theory and experiment of Fourier-Bessel field calculation and tuning of a pulsed wave annular array. *Journal of the Acoustical Society of America* 113: 2412-2423.
FUKUNAGA, K. 1990. *Introduction to Statistical Pattern Recognition*. Academic Press, New York.
GALLANT, J.L., BRAUN, J., AND VANESSEN, D.C. 1993. Selectivity for polar, hyperbolic, and Cartesian gratings in macaque visual cortex. *Science* 259, 100-103.
GALLANT, J.L., CONNOR, C.E., RAKSHIT, S., LEWIS, J.W., AND VANESSEN, D.C. 1996. Neural responses to polar, hyperbolic, and Cartesian grating in area V4 of the macaque monkey. *Journal of Neurophysiology* 76, 2718-2739.
GOMES, H.M., AND FISHER, R.B. 2003. Primal sketch feature extraction from a log-polar image. *Pattern Recognition Letters* 24, 983-992.
GROVE, T.D., AND FISHER, R.B. 1996. Attention in iconic object matching. *Proceedings of the British Machine Vision Conference* 1, Edinburgh, 293-302.
GUAN, S., LAI, C.H., AND WE, G.W. 2001. Fourier-Bessel analysis of patterns in a circular domain. *Physica D* 151: 83-98.
HAFED, Z.M., AND LEVINE, M.D. 2001. Face recognition using the discrete cosine transform. *International Journal of Computer Vision* 43: 167-188.
HARMON, L.D. 1973. The recognition of faces. *Scientific American* 229: 71-82.
HOTTA, K., KURITA, T. AND MISHIMA, T. 1998. Scale invariant face detection method using higher-order local autocorrelation features extracted from log-polar image. *Proc. Third IEEE International Conference on Automatic Face and Gesture Recognition*, 70-75.
HUBEL, D.H., AND WIESEL, T.N. 1968. Receptive field and functional architecture of monkey striate cortex. *Journal of Physiology* 195: 215-243.
ITTI, L., KOCH, C., AND BRAUN, J. 2000. Revisiting spatial vision: toward a unifying model. *Journal of the Optical Society of America A* 17, 1899-1917.
JURIE, F. 1999. A new log-polar mapping for space variant imaging: application to face detection and tracking. *Pattern Recognition* 32, 865-875.
KELLY, D.H. 1960. Stimulus pattern for visual research. *Journal of the Optical Society of America* 50, 1115-1116.
KITTLER, J., HATEF, M., DUIN, P.W., AND MATAS, J. 1998. On combining classifiers. *IEEE Transactions on Pattern Analysis and Machine Intelligence* 20: 226-239.
LADES M, VORBRUGGEN JC, BUHMANN J, LANGE J, VANDERMALSBURG C, WURTZ RP, KONEN W. 1993. Distortion invariant object recognition in the dynamic link architecture. *IEEE Transactions on Computers* 42: 300-311.
LAI, H.L., YUEN, P.C., AND FENG, G.C. 2001. Face recognition using holistic Fourier invariant features. *Pattern Recognition* 34: 95-109.
LAWRENCE, S., GILES, C.L., TSOI, A.C., AND BACK, A.D. 1997. Face recognition: a convolutional neural network approach. *IEEE Transactions on Neural Networks* 8: 98-113.
LIM, F.L., WEST, G.A.W., AND VENKATESH, S. 1997. Use of log-polar space for foveation and feature recognition. *IEE Proceedings-Vision Image and Signal Processing* 144: 323-331.
MAHON, L.E., AND DE VALOIS, R.L. 2001. Cartesian and non-Cartesian responses in LGN, V1, and V2 cells. *Visual Neuroscience* 18: 973-981.
MINUT, S., MAHADEVAN, S., HENDERSON, J., DYER, F. 2000. Face recognition using foveal vision. *Lecture Notes in Computer Science – LNCS* 1811: 424-433.
MOGHADDAM, B., JEBARA, T. AND PENTLAND, A. 2000. Bayesian face recognition. *Pattern Recognition* 33: 1771-1782.

MOVSHON, J.A., THOMPSON, I.I., AND TOLHURST, D.J. 1978. Spatial and temporal contrast sensitivity of neurons in areas 17 and 18 of the cat´s visual cortex. *Journal of Physiology* (London) 283: 79-99.
NASANEN, R. 1999. Spatial frequency bandwidth used in the face recognition of facial images. *Vision Research* 39: 3824-3833.
NASTAR, C., MOGHADDAM, B., AND PENTLAND, A. 1997. Flexible images: matching and recognition using learned deformations. *Computer Vision and Image Understanding* 65: 179-191.
OBERMAYER, K., BLASDEL, G.G., AND SCHULTEN, K. 1991. A neural network model for the formation and for the spatial structure of retinotopic maps, orientation and ocular dominance columns. In T. Kohonen, K. MÄKISARA, O. SIMULA, AND J. KANGAS, (eds.), *Artificial Neural Networks*, Elsevier, Amsterdam, Netherlands, pp. 505-511.
O´TOOLE, A.J., ABDI, H., DEFFENBACHER, K.A., VALENTIN, D. 1995. A perceptual learning theory of the information in face. In T. Valentine, (Ed.) *Cognitive and Computational Aspects of Face Recognition*. Routledge, London, pp. 159-182.
PEKALSKA, E., AND DUIN, R.P.W. 2000. Classification on dissimilarity data: a first look. *Proceedings of the 6th Annual Conference of the Advanced School for Computing and Imaging*, 221-228.
PERRET, D.I., ROLLS, E.T., AND CAAN, W. 1982. Visual neurons responsive to faces in the monkey temporal cortex. *Experimental Brain Research* 47: 329-342.
PHILLIPS, P.J., WECHSLER, H., HUANG, J., AND RAUSS, P. 1998. The FERET database and evaluation procedure for face recognition algorithms. *Image and Vision Computing Journal* 16: 295-306.
PRATT, W.K. 1991. *Digital Image Processing*. Wiley-Interscience Pub., New York.
RAUDYS, S., AND DUIN, R.P.W. 1998. On expected classification error of the Fisher linear classifier with pseudo-inverse covariance matrix. *Pattern Recognition Letters* 19: 385-392.
RAUDY, S.J., AND PIKELIS, V. 1980. On dimensionality, sample size, classification error, and complexity of classification algorithms in pattern recognition. *IEEE Transactions on Pattern Analysis and Machine Intelligence* 2: 243-251.
RHODES, G. 1988. Looking at faces: First-order and second-order features as determinants of facial appearance. Perception, 17: 43-63.
ROSENTAL, S., DAVIS, S.H., AND HOMSY, G.M. 1982. Nonlinear Marangoni convection in bounded layers. Part 1. Circular cylindrical containers. *Journal of Fluid Mechanics* 120: 91-122.
SAMARIA, F.S. 1994. *Face recognition using Hidden Markov Models*. PhD thesis, Trinity College, University of Cambridge, Cambridge.
SAMARIA, F.S., AND HARTER, A.C. 1994. Parameterization of a stochastic model for human face identification. *Proceedings of the 2nd IEEE Workshop on Applications of Computer Vision*, 138-142.
SCHWARTZ, E.L. 1977. Spatial mapping in primate sensory projection: analytic structure and relevance to perception. *Biological Cybernetics* 25: 181-194.
SCHWARTZ, E.L. 1980. Computational anatomy and functional architecture of striate cortex: a spatial mapping approach to perceptual coding. *Vision Research* 20: 645-669.
SCURICHINA, M. AND DUIN, R.P.W. 1996. Stabilizing classifiers for very small sample sizes. *Proceedings of the 13th International Conference on Pattern Recognition* 2, Track B: 891-896.
SERGENT, J. 1986. Microgenesis of face perception. In *Aspects of Face Processing.*, H.D. ELLIS, M.A, JEEVES, F. NEWCOMBE, AND A. YOUNG, Eds. Nijhoff, Dordrecht, The Netherlands.
SHEPARD, R. N. AND CERMAK, G. W. 1973. Perceptual-cognitive explorations of a toroidal set of free-form stimuli. *Cognitive Psychology* 4: 351-377.
SHEPARD, R. N. AND CHIPMAN, S. 1970. Second-order isomorphism of internal representations: Shapes of states. *Cognitive Psychology* 1: 1-17.
SIM, T., SUKTHANKAR, R., MULLIN, M., AND BALUJA, S. 2000. Memory-based recognition for visitor identification. *Proceedings of the 4th International Conference on Automatic Face and Gesture Recognition:* 214-220.
SMERALDI, F. AND J. BIGUN, J. 2002. Retinal vision applied to facial features detection and face authentication. *Pattern Recognition Letters* 23: 463 - 475.
SPANIER, J. AND OLDHAM, K.B. 1987. *An Atlas of Functions*. Hemisphere Publishing Corporation, Washington, DC
TIEGER, T., AND GANZ, L. 1979. Recognition of faces in the presence of two-dimensional sinusoidal masks. *Perception and Psychophysics* 26: 163-167.
TISTARELLI, M., AND GROSSO, E. 1998. Active vision-based face recognition issues, applications and techniques. In Wechsler, H., et al. (Eds.), *Nato-Asi Advanced Study on Face Recogniton*, Vol. F-163. Springer, Berlin, pp. 262-286.
TURK, M., AND PENTLAND, A. 1991. Eigenfaces for recognition. *Journal of Cognitive Neuroscience* 3: 71-86.
WILSON, H.R. 1991. Psychophysical models of spatial vision and hyperacuity. In D. Regan (Ed.), *Spatial Vision*. Macmillan: London, pp. 64-86.
WILSON, H.R, WILKINSON, F. AND ASAAD, W. 1997. Concentric orientation summation in human form vision. *Vision Research* 37: 2325-2330.
WILSON, H.R., AND WILKINSON, F. 1998. Detection of global structure in Glass patterns: implications for form vision. *Vision Research* 38: 2933-2947.

WILSON, H.R., LOFFLER, G., WILKINSON, F. AND THISTLETHWAITE, W.A. 2001. An inverse oblique effect in human vision. *Vision Research* 41: 1749-1753.
WISKOTT, L., FELLOUS, J., KRÜGER, N., AND VON DER MALSBNURG, C. 1997. Face recognition by elastic bunch graph matching. *IEEE Transactions on Pattern analysis and Machine Intelligence* 19: 775-779.
YOUNG, M.P., AND YAMANE, S. 1992. Sparse population coding of faces in the inferotemporal cortex. *Science* 256: 1327-1331.
ZWICK, M., AND ZEITLER, E. 1973. Image reconstruction from projections. *Optik* 38, 550-565.